\documentclass[journal]{IEEEtran}

\usepackage{amsmath,amssymb,amsfonts}
\usepackage{algorithmic}
\usepackage{algorithm}
\usepackage{graphicx}
\usepackage{textcomp}
\usepackage{xcolor}
\usepackage{cite}
\usepackage{booktabs}
\usepackage{multirow}
\usepackage{makecell}  % For header line breaks
\usepackage{url}
\usepackage{placeins}  % For \FloatBarrier to control float placement

\usepackage[hidelinks]{hyperref}

\graphicspath{{assets/figures/}}

\begin{document}

% ============================================================
% Title
% ============================================================
\title{AnchorDrive: LLM Scenario Rollout with Anchor-Guided Diffusion Regeneration for Safety-Critical Scenario Generation}

% ============================================================
% Authors
% ============================================================
\author{Zhulin~Jiang$^{*}$,~Zetao~Li$^{*}$,~Cheng~Wang,~Ziwen~Wang,~Chen~Xiong$^{\dagger}$%
\thanks{$^{*}$These authors contributed equally to this work.}%
\thanks{$^{\dagger}$Corresponding author (e-mail: xiongch8@mail.sysu.edu.cn).}%
\thanks{All authors are with the School of Intelligent Systems Engineering, Sun Yat-sen University, Shenzhen, China.}%
\thanks{The authors would like to thank the fundings of the National Key R\&D Program of China (Grant No. 2024YFB4303400), Natural Science Foundation of Guangdong Province, China (Grant No. 2025A1515010166) and the Shenzhen Fundamental Research Program, China (Grant No. JCYJ20240813151301003).}%
}

% Header
\markboth{IEEE Transactions on Vehicular Technology}%
{Jiang \MakeLowercase{\textit{et al.}}: AnchorDrive}

\maketitle

% ============================================================
% Abstract
% ============================================================
\begin{abstract}
Autonomous driving systems require comprehensive evaluation in safety-critical scenarios to ensure safety and robustness. However, such scenarios are rare and difficult to collect from real-world driving data, necessitating simulation-based synthesis. Yet, existing methods often exhibit limitations in both controllability and realism. From a capability perspective, LLMs excel at controllable generation guided by natural language instructions, while diffusion models are better suited for producing trajectories consistent with realistic driving distributions. Leveraging their complementary strengths, we propose AnchorDrive, a two-stage safety-critical scenario generation framework. In the first stage, we deploy an LLM as a driver agent within a closed-loop simulation, which reasons and iteratively outputs control commands under natural language constraints; a plan assessor reviews these commands and provides corrective feedback, enabling semantically controllable scenario generation. In the second stage, the LLM extracts key anchor points from the first-stage trajectories as guidance objectives, which jointly with other guidance terms steer the diffusion model to regenerate complete trajectories with improved realism while preserving user-specified intent. Experiments on the highD dataset demonstrate that AnchorDrive achieves superior overall performance in criticality, realism, and controllability, validating its effectiveness for generating controllable and realistic safety-critical scenarios.
\end{abstract}

% ============================================================
% Keywords
% ============================================================

\IEEEpeerreviewmaketitle

% ============================================================
% 1. Introduction
% ============================================================
\section{Introduction}

As autonomous driving technology continues to advance, validating the safety and robustness of autonomous driving systems across diverse scenarios has become an indispensable step toward practical deployment~\cite{10591480,10716596,8667012}. Testing in safety-critical scenarios is particularly essential, as such scenarios serve as the core basis for assessing system safety capabilities. These scenarios typically exhibit high-risk interaction characteristics: autonomous vehicles may encounter sudden or unconventional behaviors from other road users during interactions, significantly elevating the likelihood of collisions~\cite{10089194,wang2025risk}. However, real-world road testing is not only resource-intensive but also makes it difficult to systematically collect sufficient and adequately dangerous safety-critical samples, resulting in limited evaluation coverage. In contrast, simulation-based safety-critical scenario generation and testing has become a prevalent approach due to its lower cost, higher efficiency, and ease of scalability and reproducibility.

\begin{figure}[!t]
\centering
\includegraphics[width=\linewidth]{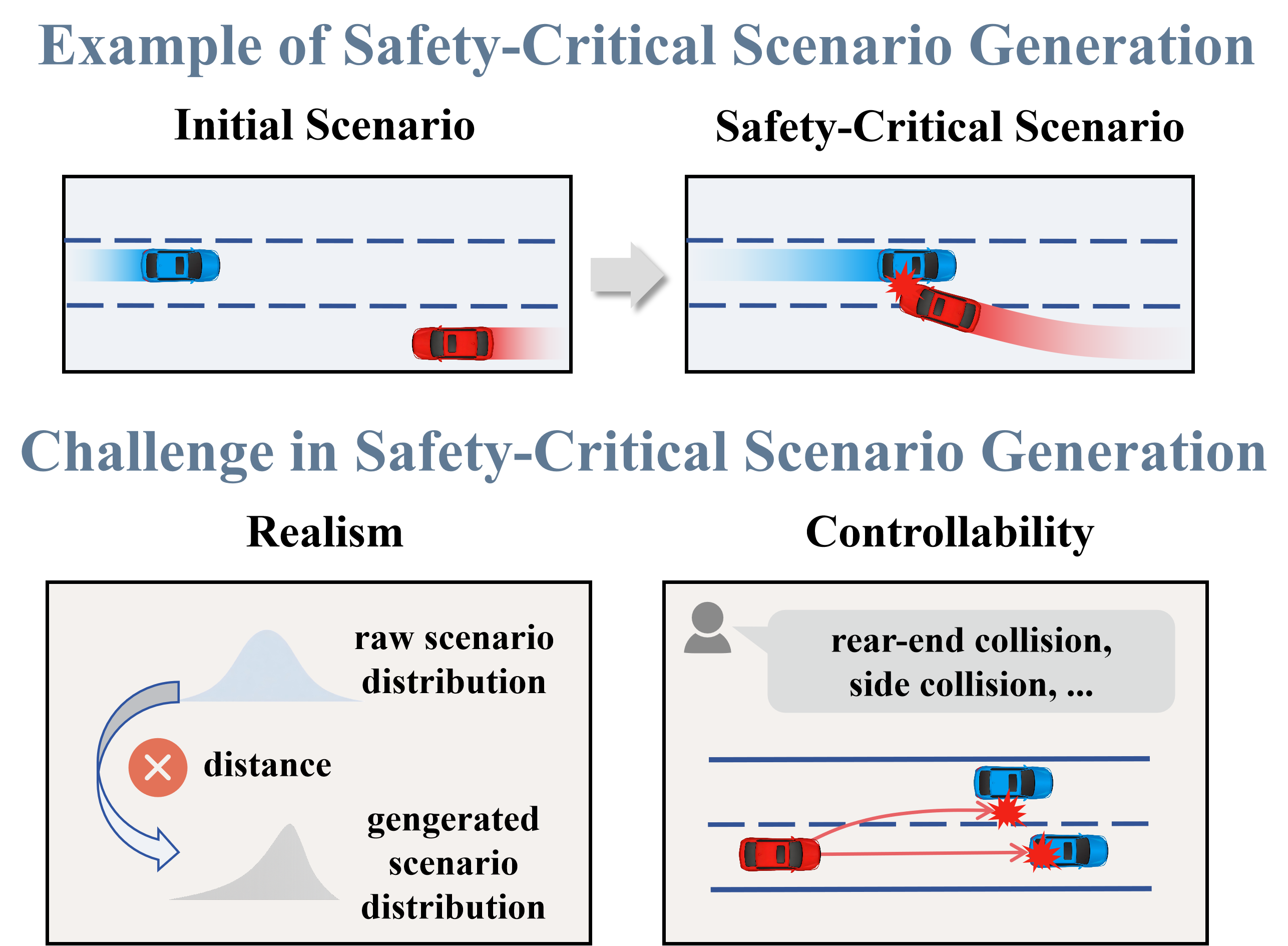}
\caption{Illustrative example and main challenges of scenario generation. The blue vehicle represents the ego vehicle, while the red vehicle represents the adversarial vehicle.}
\label{fig:scenario_challenges}
\end{figure}

As illustrated in Fig.~\ref{fig:scenario_challenges}, methods for generating safety-critical scenarios in simulation typically configure road structures, traffic participants, their initial states, and behavioral policies within virtual environments, synthesizing trajectories that trigger high-risk interactions for safety evaluation of the system under test. Such generation requires the results to possess both realism and controllability. Realism requires that kinematic characteristics of generated trajectories (e.g., velocity, acceleration) statistically approximate real driving data distributions as closely as possible, while avoiding unreasonable behaviors (e.g., lane departure), thereby preventing extreme and implausible scenarios. Controllability requires that generated results adhere to user semantic descriptions, enabling targeted generation of specified risk interaction scenarios. However, existing methods often struggle to satisfy both realism and controllability.

Early safety-critical scenario generation methods typically relied on simulation platforms such as CARLA~\cite{dosovitskiy2017carla} and SUMO~\cite{lopez2018microscopic}, where researchers manually or semi-automatically designed scenarios based on ontologies~\cite{da2024ontology,scanlon2021waymo}. Such methods offer strong controllability but suffer from low efficiency, heavy reliance on expert knowledge, and insufficient realism. To improve the realism of generated scenarios, deep learning-based methods have emerged. These methods typically train models on large-scale real driving data and generate safety-critical scenarios at test time through optimization-based techniques or gradient guidance~\cite{jiang2023motiondiffuser,xu2025diffscene,rempe2022generating,wang2021advsim}. Since deep learning models can capture behavioral patterns present in the training distribution, such methods exhibit clear advantages in scenario realism; however, they often struggle to finely control generation results, exhibiting limitations in controllability.

The advancement of generative AI offers new possibilities for addressing these challenges. LLMs have demonstrated exceptional capabilities in natural language understanding and code generation~\cite{achiam2023gpt,bubeck2023sparks,chen2021evaluating}, making them naturally suited as interactive interfaces for users to describe safety-critical scenarios via natural language. Works such as ChatScene and OmniTester compile user descriptions into scripts or configuration files to drive simulators for scenario generation~\cite{zhang2024chatscene,lu2025omnitester}, achieving a text-to-scenario paradigm that significantly lowers the barrier for scenario construction; however, generated trajectories still exhibit considerable gaps from real trajectory distributions. Meanwhile, LLMs have also been employed to translate semantic descriptions into loss terms for guiding diffusion model denoising~\cite{zhong2023language,peng2025ld}, enhancing controllability of deep learning-based methods to some extent. However, the mapping from natural language to loss functions involves representational mismatches, and complex interactions are difficult to stably encode as loss functions, resulting in responses to instructions that remain insufficiently direct and fine-grained, with room for further improvement.

Existing methods often generate scenarios in a single pipeline, which tends to result in insufficient realism or limited controllability. To address this, we decouple these two aspects: first solving the controllable generation problem, then addressing the realism problem. We draw inspiration from two recent lines of progress: end-to-end autonomous driving research has shown that LLM-based driver agents with ``perception--reasoning--execution'' closed-loop capabilities can stably translate high-level intentions into executable action sequences~\cite{fu2024drive,zhou2025opendrivevla,zhou2025autovla,shao2024lmdrive}, enabling their adaptation as interaction process constructors for generating high-risk trajectories consistent with user instructions. Meanwhile, diffusion models have demonstrated strong capabilities in fitting real data distributions for trajectory generation, and can guide the denoising process through anchor points to ensure generated trajectories satisfy specified conditions at critical moments and positions~\cite{huang2024versatile,chang2024safe}, thereby producing trajectories that both pass through designated anchors and remain realistic. Based on this complementarity, we first employ an LLM-based driver agent to generate controllable scenarios, then use anchor-guided diffusion for trajectory regeneration, enabling results that achieve both controllability and realism.

Therefore, we propose AnchorDrive, a two-stage LLM-guided safety-critical scenario generation framework. Our framework comprises two main modules: an LLM-based Scenario Rollout module for controllable scenario generation, and a trajectory regeneration module for optimization. In the first module, the LLM-based driver agent receives environment states and user natural language instructions, performs reasoning under preset Chain-of-Thought (CoT) prompts, and iteratively outputs control commands; a plan assessor reviews these commands and provides corrective feedback upon failures, suppressing unreasonable behaviors to stably generate safety-critical scenarios conforming to user instructions within the simulation closed-loop. In the second module, we utilize the LLM to analyze first-stage trajectories and automatically extract key anchor points, constructing these anchors along with other constraints as guidance objectives to form a loss function that guides the diffusion model's denoising process for regenerating complete multi-vehicle trajectories. The anchors preserve driving intentions and temporal structure of critical events from the first stage, while the diffusion prior enhances kinematic plausibility and statistical consistency, thereby improving scenario realism without altering scenario semantics (risk pattern type, key participants, and temporal ordering of critical events).

In summary, the main contributions of this paper include:
\begin{itemize}
\item We propose a safety-critical scenario generation method based on a closed-loop LLM-based driver agent: iteratively outputting control sequences under natural language constraints, with review and corrective feedback from a plan assessor to suppress unreasonable behaviors, thereby stably generating high-risk interaction scenarios consistent with instruction semantics and achieving semantically controllable generation.

\item We design a diffusion model-based trajectory optimization mechanism: the LLM extracts a sparse set of key anchor points from generated trajectories, using these anchors along with other guidance objectives to steer diffusion denoising for multi-vehicle trajectory regeneration and refinement, improving trajectory realism while preserving scenario semantics.

\item We propose the two-stage AnchorDrive framework and conduct systematic evaluation on the highD dataset. Experiments demonstrate superior overall performance across criticality, realism, and controllability metrics, validating the effectiveness for generating controllable and realistic safety-critical scenarios.
\end{itemize}

The remainder of this paper is organized as follows: Section~II reviews related work on safety-critical scenario generation; Section~III presents the proposed AnchorDrive framework; Section~IV provides experimental evaluation and component effectiveness analysis on highD; Section~V concludes the paper and discusses future directions.

% ============================================================
% 2. Related Work
% ============================================================
\section{Related Work}

\subsection{Traditional Scenario Generation Methods}

Early scenario generation methods primarily relied on experts designing safety-critical scenarios within simulation platforms such as CARLA and SUMO, manually adjusting participants' initial positions, velocities, and driving routes, or constructing collisions, near-misses, and traffic violations using ontologies and script templates~\cite{da2024ontology,scanlon2021waymo}. Such methods offer strong controllability at the parameter level but suffer from heavy reliance on domain knowledge, limited scalability, and difficulty maintaining consistency with real trajectory distributions~\cite{zhang2022rethinking}.

Recent research has explored specific parameterized spaces within original scenarios to precisely locate adversarial parameters using optimization-based methods~\cite{abeysirigoonawardena2019generating,ding2020cmts,hanselmann2022king}. For example, AdvSim~\cite{wang2021advsim} directly perturbs background vehicles in trajectory space to search for trajectories that cause planner failures. Strive~\cite{rempe2022generating} introduces deep learning models to optimize in data-driven latent spaces while enhancing multi-vehicle interaction plausibility; other works train adversarial vehicles via reinforcement learning to learn dangerous maneuvers in closed-loop interactions~\cite{ding2020learning,koren2019efficient}. Although these methods can systematically discover challenging scenarios, the generated results are often determined by optimization objectives or reward designs, making it difficult to specify risk patterns at a high-level semantic layer, thus generally lacking semantic-level controllability.

In recent years, guidance-based diffusion models have been introduced to scenario generation to improve controllability of deep learning-based methods. Works such as CTG guide the denoising sampling process through differentiable loss functions, STL constraints, or safety objectives for controllable trajectory generation~\cite{jiang2023motiondiffuser,xu2025diffscene,rempe2022generating}; SafeSim~\cite{chang2024safe} further combines loss function guidance with partial diffusion in closed-loop simulation, enabling adjustment of collision types and risk levels while maintaining behavioral realism. However, such methods still heavily rely on manually designed loss functions or task-specific reward models; different adversarial objectives often require reconstructing loss functions, limiting flexible specification at the semantic level and reducing usability for non-expert users.

\subsection{LLM-based Scenario Generation}

Benefiting from strong capabilities in natural language understanding, knowledge reasoning, and code generation, LLMs have been widely applied to scenario generation tasks in recent years~\cite{gao2025foundation,chang2024llmscenario}, demonstrating potential for generating realistic and controllable scenarios.

One class of methods utilizes LLMs to compile user descriptions into intermediate representations such as scripts or configurations, defining participant initial states, interaction rules, and trigger conditions for rapid construction and batch generation. For example, LCTGen~\cite{tan2023language} uses LLMs to convert text queries into structured representations, then employs Transformers to generate traffic scenarios. ChatScene~\cite{zhang2024chatscene} analyzes user descriptions via LLMs, translating natural language into Scenic scripts and executing them in CARLA to obtain corresponding scenarios. These methods significantly lower the barrier for scenario generation, allowing users to directly obtain target dangerous scenarios through simple descriptions; however, they still face the problem of insufficient scenario realism.

Another class of methods combines LLMs with diffusion models. CTG++~\cite{zhong2023language} uses LLMs to convert user natural language instructions into differentiable guidance loss functions, applying gradient guidance during diffusion model denoising to ensure generated trajectories satisfy semantic constraints described in instructions. LD-Scene~\cite{peng2025ld} applies this approach to safety-critical scenario generation, using chain-of-thought prompts to generate adversarial loss functions that guide latent diffusion models to generate specified types of safety-critical scenarios. However, the mapping from natural language to loss functions remains insufficiently direct, and complex interaction intents are difficult to stably encode as optimization objectives, thus limiting generation controllability.

\subsection{LLM-based Driver Agents}

Recent end-to-end autonomous driving research has begun deploying LLMs directly as driver agents, generating driving decisions in closed-loop within simulation environments~\cite{mao2023language,wen2023dilu,hu2024agentscodriver}. \cite{li2024large} notes that LLMs can serve as high-level decision-makers, driving stably on roads.

In early work, Fu et al.~\cite{fu2024drive} constructed an LLM-driven closed-loop control framework in a simplified highway environment: encoding surrounding information as concise text input to the LLM for discrete action selection with explanations. LMDrive~\cite{shao2024lmdrive} encodes front-view camera images with navigation text and inputs them to the LLM, which directly outputs steering angles and acceleration to control the vehicle, achieving navigation-text-guided end-to-end closed-loop control. DriveGPT4~\cite{xu2024drivegpt4} extends this paradigm to more open urban traffic, driving step-by-step reasoning through semantic summaries and outputting continuous control signals, demonstrating LLMs' capabilities for long-horizon planning and complex interaction modeling.

Overall, LLM-based driver agents possess the advantage of following natural language intent and generating interpretable driving behaviors in closed-loop interactions. Based on this, we adopt an LLM-based driver agent in the first stage to construct safety-critical scenarios consistent with user descriptions.

% ============================================================
% 3. Methodology
% ============================================================
\section{Methodology}

This section presents the technical details of the proposed AnchorDrive framework. We first provide a formal definition of the safety-critical scenario generation task, clarifying inputs, outputs, and notation; then we describe the specific implementations of the first and second stages respectively.

\subsection{Problem Formulation}

Our objective is to generate controllable and realistic safety-critical interaction scenarios conditioned on natural language instructions. Each scenario contains $N$ traffic participants, where the ego vehicle and adversarial vehicle are specified by user instruction $I$. We aim to make the adversarial vehicle collide with the ego vehicle in a user-specified manner while ensuring all vehicle trajectories remain realistic.

A traffic scenario consists of a map $M$ and state sequences of $N$ traffic participants, where the map contains lane topology, drivable areas, etc. We index traffic participants by $i \in \{0, 1, \ldots, N-1\}$. At a discrete time step $t$ (with time step length set by the simulator), the joint state of all participants is denoted as $s_t = [s_t^0, s_t^1, \ldots, s_t^{N-1}]$. The single-vehicle state is defined as $s_t^i = (x_t^i, y_t^i, \theta_t^i, v_t^i)$, where $x_t^i, y_t^i, \theta_t^i, v_t^i$ represent 2D position coordinates, heading angle, and velocity respectively. Given a history observation window of length $T_{\mathrm{hist}}$, we denote the history state sequence as $x = \{s_{t-T_{\mathrm{hist}}}, s_{t-T_{\mathrm{hist}}+1}, \ldots, s_t\}$, i.e., historical states of all traffic participants over the past $T_{\mathrm{hist}}$ time steps. We generate joint trajectories $\tau$ of all participants over the future prediction horizon $[t+1, t+T]$, consisting of action sequence $\tau^a$ and state sequence $\tau^s$: $\tau = [\tau^a, \tau^s]$, where $\tau^s = \{s_{t+1}, s_{t+2}, \ldots, s_{t+T}\}$ and $\tau^a = \{a_{t+1}, a_{t+2}, \ldots, a_{t+T}\}$. At time $t$, the joint action is defined as $a_t = [a_t^0, a_t^1, \ldots, a_t^{N-1}]$, where single-vehicle action $a_t^i = (\dot{v}_t^i, \dot{\theta}_t^i)$, with $\dot{v}_t^i$ and $\dot{\theta}_t^i$ representing acceleration and yaw rate respectively. The action sequence $\tau^a$ is integrated via vehicle kinematics to obtain state sequence $\tau^s$, thus determining future trajectory $\tau$; therefore, our task is essentially to predict action sequence $\tau^a$.

User instruction $I$ specifies the desired risk interaction pattern to generate (e.g., ``adversarial vehicle cuts in causing rear-end collision''), which may include target participants, interaction manner, and expected risk outcome. Given inputs $(M, x, I)$, we aim to generate future trajectory $\tau$ satisfying two requirements: 1) Controllability: the generated interaction process is consistent with user instruction description, with adversarial vehicle colliding with ego vehicle in the user-specified manner; 2) Realism: kinematic characteristics of generated trajectories statistically approximate real driving data distributions as closely as possible, while avoiding unreasonable behaviors. Therefore, we formulate the problem as learning a conditional generative model
\begin{IEEEeqnarray}{c}
\tau \sim P_\theta(\tau \mid M, x, I)
\end{IEEEeqnarray}
for directly generating multi-vehicle future joint trajectories $\tau$ conditioned on map $M$, historical observations $x$, and natural language instruction $I$, yielding safety-critical interaction scenarios with both controllability and realism.

\begin{figure*}[!t]
\centering
\includegraphics[width=\textwidth]{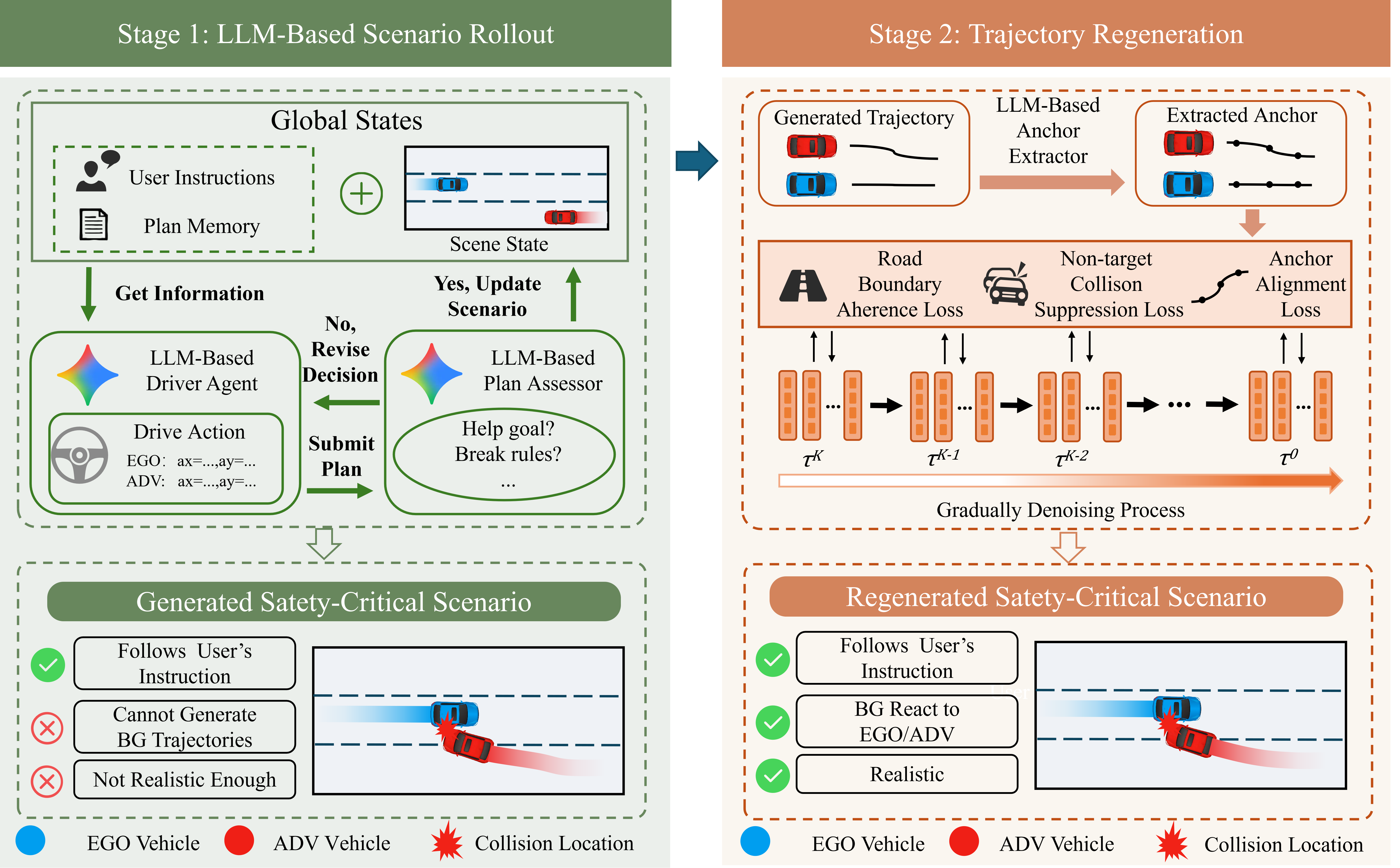}
\caption{Overall architecture of the AnchorDrive framework. The framework employs two-stage generation: Stage 1 generates safety-critical scenarios satisfying user instructions in a closed-loop manner under natural language instruction constraints; Stage 2 extracts a sparse set of key anchor points from Stage 1 trajectories and jointly guides the diffusion model denoising process with other constraints to regenerate complete trajectories, improving multi-vehicle trajectory realism while preserving scenario semantics.}
\label{fig:framework}
\end{figure*}

\subsection{Stage 1: LLM-based Scenario Rollout}

This section describes the closed-loop generation process of the first stage (LLM-based Scenario Rollout). As shown in Fig.~\ref{fig:framework}, the generation process iterates with fixed-length frame windows as planning steps. At each planning step, the LLM-based driver agent reads global state and, after reasoning about the current situation, outputs the plan for that step, including driving intentions and corresponding per-frame control commands for both ego and adversarial vehicles. For convenient direct vehicle control, we represent each frame's control command uniformly as $(a_x, a_y)$, representing lateral and longitudinal acceleration respectively. Subsequently, commands are sent to the LLM-based plan assessor for rapid review, focusing on ``whether goal is advanced'' and ``whether rule constraints are violated.'' If passed, the command is executed to update the scenario, and the planning decision is written to plan memory; if not passed, it returns to the LLM-based driver agent for plan revision. This closed-loop iterates by time steps until the goal is achieved or maximum simulation steps are reached. Specifically, this module comprises three submodules:

\subsubsection{Global State Constructor}

The global state constructor provides the LLM-based driver agent with global information required for decision-making, comprising three parts: user instruction, scenario state, and planning memory. User instruction is the user's natural language description of the desired scenario, e.g., ``adversarial vehicle changes lanes to cut in front of ego vehicle causing rear-end collision.'' Scenario state depicts current traffic conditions, including positions, heading angles, velocities, dimensions, and types of ego vehicle, adversarial vehicle, and background vehicles within a certain range, as well as road geometry information such as lane centerlines, road boundaries, and adjacent lane topology; scenario screenshots are also provided to enhance LLM understanding of spatial relationships. Planning memory records summaries of executed planning steps during closed-loop iteration, including driving intentions, per-frame control commands, and executed trajectories for each step, serving as references for subsequent reasoning.

\subsubsection{LLM-based Driver Agent}

The LLM-based driver agent reads global state and generates driving decisions under prompt constraints. LLM reasoning effectiveness is significantly influenced by prompt design~\cite{liu2023pre,sahoo2024systematic}; to improve reasoning stability, we design hierarchical CoT prompts that guide each planning step through a ``situation analysis--intention planning--action output'' workflow. The situation analysis layer evaluates relative position and velocity relationships between ego and adversarial vehicles, as well as spatial constraints with background vehicles and lane boundaries, thereby constraining the feasible action space and avoiding boundary violations and non-target collisions. The intention planning layer generates high-level driving intentions for ego and adversarial vehicles based on user instruction and current state. For example, for instruction ``adversarial vehicle changes lanes to cut in front of ego vehicle causing rear-end collision,'' if the adversarial vehicle has entered the lane-change process, this layer requires it to continue completing the cut-in and decelerate appropriately after cutting in to promote rear-end collision, while the ego vehicle maintains original speed. The action output layer outputs driving intentions and per-frame control sequences for ego and adversarial vehicles following a template, with each frame as $(a_x, a_y)$, constraining value ranges to avoid extreme actions.

\subsubsection{LLM-based Plan Assessor}

We introduce the LLM-based plan assessor to review outputs from the LLM-based driver agent. It integrates per-frame controls through vehicle kinematics into short-term trajectories, with background vehicles following original dataset trajectories, and evaluates from two aspects: goal advancement (whether interaction conforms to instruction) and constraint satisfaction (whether boundary violations occur, whether collisions with background vehicles occur, whether control magnitudes are feasible). If evaluation passes, controls are executed in the simulation environment to update scenario state, and driving intentions with per-frame controls for that planning step are written to planning memory; if not passed, structured failure reasons are returned, driving the LLM-based driver agent to regenerate until passing or reaching maximum retry count.

Through multiple closed-loop iterations, the first stage can generate safety-critical scenarios consistent with user instructions; however, its output still has two limitations: First, background vehicles lack interactive response---this stage only plans ego and adversarial vehicles, while other background vehicles follow original dataset trajectories and cannot react to main vehicle behaviors; Second, trajectory realism is insufficient---the LLM outputs control commands as per-frame accelerations, easily introducing unnatural motions such as velocity and acceleration discontinuities, and statistical distributions of key kinematic metrics also differ significantly from the original dataset.

\subsection{Stage 2: Diffusion Model Trajectory Regeneration for Generated Scenario Optimization}

To optimize scenarios generated in the first stage, we design a diffusion model-based scenario optimization mechanism that employs diffusion models to regenerate first-stage trajectories for improved scenario realism. As shown in Fig.~\ref{fig:framework}, the second stage comprises two submodules: LLM-based anchor extractor and anchor-guided diffusion trajectory regeneration module. The LLM-based anchor extractor performs semantic and kinematic analysis on first-stage trajectories, extracting key anchor points as constraints for preserving scenario semantics. The anchor-guided diffusion trajectory regeneration module combines anchor alignment, road boundary adherence, and non-target collision suppression as three guidance objectives into a loss function, applying gradient guidance to the diffusion model during denoising to make regenerated trajectories more realistic while maintaining critical interaction structures.

\subsubsection{LLM-based Anchor Extractor}

Anchors serve as the critical bridge connecting the two stages. First-stage trajectories contain per-frame state information; if all frames were used as guidance targets for the diffusion model, it would significantly compress adjustable degrees of freedom, preventing the model from leveraging its advantages in detail refinement and distribution fitting; it would also limit correction of unreasonable behaviors---for example, once a trajectory exhibits boundary violations, per-frame guidance would force regenerated results to follow original deviations, making effective correction difficult. Therefore, we design an LLM-based anchor extractor that identifies and extracts an appropriate number of key anchor points from complete trajectories (such as intention initiation, interaction occurrence, collision occurrence, etc.), applying guidance only at critical moments while leaving other moments for the diffusion model to complete and refine, thereby improving trajectory realism while preserving interaction semantics.

The anchor extractor receives complete records output from the first stage, including driving intention descriptions and per-frame control commands for each planning step along with corresponding trajectories. Through prompt guidance, the LLM analyzes trajectories, identifies key phases of ego and adversarial vehicle behaviors, and selects representative spatiotemporal positions $(x, y, t)$ as anchor points for each phase. Anchors need not exactly replicate precise coordinates from original trajectories; appropriate adjustments are permitted for better generation results.

Since safety-critical scenarios involve adversarial vehicles attacking ego vehicles, for adversarial vehicles, we further attach brief explanations to anchors describing the behavioral phase or intention node at that anchor. Meanwhile, to maintain temporal consistency of two-vehicle interactions, we synchronously set corresponding anchors at the same frame numbers in ego vehicle trajectories without requiring semantic explanations.

\subsubsection{Anchor and Rule Consistency Joint-Guided Diffusion Trajectory Regeneration}

To make trajectories more realistic while preserving first-stage scenario semantics, we propose a multi-objective guided diffusion model trajectory regeneration strategy for optimizing already-generated scenarios. This strategy uses LLM-extracted key anchors as primary guidance objectives, and unifies road boundary adherence and non-target collision suppression into composable guidance loss functions, applying gradient corrections on clean trajectory estimates at each denoising step to stably achieve semantic preservation and realism improvement.

Specifically, following~\cite{zhong2022guided}, our diffusion model directly predicts action sequence $\tau^a$, while state sequence $\tau^s$ can be derived from initial state $s_t$ and dynamics model $f$. The diffusion model generates trajectories by reversing a gradual noising process. Given real trajectory $\tau_0$ sampled from data distribution $q(\tau_0)$, the forward noising process yields a sequence of increasingly noisy trajectories $(\tau_1, \tau_2, \ldots, \tau_K)$. The $k$-th step trajectory $\tau_k$ is obtained by adding Gaussian noise to $\tau_{k-1}$, parameterized by predefined variance schedule $\beta_k$:
\begin{IEEEeqnarray}{c}
q(\tau_{1:K}|\tau_0) := \prod_{k=1}^{K} q(\tau_k|\tau_{k-1})
\end{IEEEeqnarray}
\begin{IEEEeqnarray}{c}
q(\tau_k|\tau_{k-1}) := \mathcal{N}(\tau_k; \sqrt{1-\beta_k}\tau_{k-1}, \beta_k I)
\end{IEEEeqnarray}

The noising process gradually corrupts data, making final $q(\tau_K)$ approach $\mathcal{N}(\tau_K; 0, I)$. Trajectory generation is achieved by learning the reverse of this noising process: given noisy trajectory $\tau_K$, the model learns to denoise it back to $\tau_0$ through a series of reverse steps. Each reverse step is modeled as:
\begin{IEEEeqnarray}{c}
p_\theta(\tau_{k-1}|\tau_k, c) := \mathcal{N}(\tau_{k-1}; \mu_\theta(\tau_k, k, c), \Sigma_k)
\end{IEEEeqnarray}
where $c$ is conditioning information including map road structure $M$ and all vehicle history trajectories $x$, and $\Sigma_k$ uses preset variance schedule. Specifically, we employ noise prediction parameterization with network output $\hat{\epsilon}_\theta = \epsilon_\theta(\tau_k, k, c)$, and use the closed-form relationship from the forward process to recover the clean trajectory from $\tau_k$, obtaining current step's $\hat{\tau}_0$ estimate:
\begin{IEEEeqnarray}{c}
\hat{\tau}_0 = \frac{\tau_k - \sqrt{1-\bar{\alpha}_k} \hat{\epsilon}_\theta}{\sqrt{\bar{\alpha}_k}}
\end{IEEEeqnarray}
where $\bar{\alpha}_k$ is a DDPM predefined parameter. Subsequently, we fuse $\hat{\tau}_0$ with $\tau_k$ according to DDPM posterior mean form to compute the reverse step mean:
\begin{IEEEeqnarray}{rCl}
\mu_\theta(\tau_k, k, c) &=& \frac{\sqrt{\bar{\alpha}_{k-1}}\beta_k}{1-\bar{\alpha}_k}\hat{\tau}_0 \nonumber\\
&& {} + \frac{\sqrt{\alpha_k}(1-\bar{\alpha}_{k-1})}{1-\bar{\alpha}_k}\tau_k
\end{IEEEeqnarray}
and sample $\tau_{k-1}$ from this Gaussian distribution; repeating this iteration yields final generated trajectory $\tau_0$.

While diffusion priors alone can generate realistic behaviors, they cannot satisfy user instructions. Therefore, we employ the reconstruction guidance method proposed in~\cite{ho2022video,rempe2023trace} to guide each denoising step of the diffusion model: at the $k$-th reverse denoising step, the diffusion model first reconstructs the corresponding clean trajectory estimate $\hat{\tau}_0$ from the current noisy sample. Subsequently, we compute the gradient of loss function $J$ on this clean trajectory estimate $\hat{\tau}_0$ and apply a small correction to $\hat{\tau}_0$ along the gradient direction:
\begin{IEEEeqnarray}{c}
\tilde{\tau}_0 = \hat{\tau}_0 - \alpha \Sigma_k \nabla_{\tau_k} J(\hat{\tau}_0)
\end{IEEEeqnarray}
where $\alpha$ is guidance strength controlling the magnitude of applied guidance. This strategy stably injects guidance objective information into each denoising step, making the corrected $\tilde{\tau}_0$ more inclined toward satisfying guidance objectives, thereby enabling generated trajectory $\tau_0$ to better satisfy various guidance objectives.

Specifically, we write guidance objectives as composable weighted loss functions:
\begin{IEEEeqnarray}{rCl}
J(\tau) &=& \lambda_{\mathrm{anchor}} J_{\mathrm{anchor}} + \lambda_{\mathrm{avoid}} J_{\mathrm{avoid}} \nonumber\\
&& {} + \lambda_{\mathrm{boundary}} J_{\mathrm{boundary}}
\end{IEEEeqnarray}
where $\lambda_{\mathrm{anchor}}, \lambda_{\mathrm{avoid}}, \lambda_{\mathrm{boundary}}$ control relative strengths of each guidance objective. The three loss terms correspond to: key semantic anchor alignment, non-target collision suppression, and road boundary adherence loss.

\textbf{a) Anchor Alignment Loss $J_{\mathrm{anchor}}$}: To preserve critical interaction structures determined in the first stage, formally, the anchor extractor maps first-stage records $R$ to anchor set $A$, denoted as $A = \operatorname{E}_{\mathrm{LLM}}(R)$, where $A$ can be expressed as $A = \{(p_i, t_i)\}_{i=1}^{N_a}$ with $p_i = (x_i, y_i)$. Let the position of diffusion model clean trajectory estimate $\hat{\tau}_0$ at time $t_i$ be $p(t_i) = (x(t_i), y(t_i))$, then
\begin{IEEEeqnarray}{c}
J_{\mathrm{anchor}} = \sum_{i=1}^{N_a} \| p(t_i) - p_i \|_2^2
\end{IEEEeqnarray}
where $N_a$ is total anchor count. This term encourages vehicles to reach corresponding anchor positions at time $t_i$, thereby maintaining critical interaction semantic consistency.

\textbf{b) Non-target Collision Suppression Loss $J_{\mathrm{avoid}}$}: In safety-critical scenarios, we wish to preserve only the intended target collision pair, i.e., ego and adversarial vehicles collide, while non-target vehicle pairs should maintain safe distances. Let the bounding box gap between vehicles $i$ and $j$ at time $t$ be $d_{ij}(t)$, safety distance threshold be $d_{\mathrm{safe}}$, and denote the set of vehicle pairs requiring guidance as $B$, containing pairs of ego-background, adversarial-background, and background-background vehicles. The loss function is:
\begin{IEEEeqnarray}{rCl}
J_{\mathrm{avoid}} &=& \sum_{t=1}^{T} \sum_{(i,j) \in B} \max(0, d_{\mathrm{safe}} - d_{ij}(t)) \nonumber\\
&& {} \cdot \exp\left(\frac{d_{\mathrm{safe}} - d_{ij}(t)}{\sigma}\right)
\end{IEEEeqnarray}
where $\sigma$ is penalty strength parameter. When inter-vehicle distance falls below safety threshold, this term increases rapidly with penetration depth, effectively suppressing non-target collisions.

\textbf{c) Road Boundary Adherence Loss $J_{\mathrm{boundary}}$}: Used to prevent vehicles from entering non-drivable areas. Since highD is a highway dataset, the main non-drivable areas are reflected in lateral boundary constraints, so we apply boundary penalties only to vehicle lateral positions. Let vehicle $i$'s lateral coordinate at time $t$ be $y_i(t)$. We define each vehicle $i$'s corresponding lateral drivable interval $[y_{\mathrm{lower}}^{(i)}, y_{\mathrm{upper}}^{(i)}]$ and introduce safety margin $\delta_{\mathrm{boundary}}$. When vehicles approach or exceed their drivable boundaries, loss increases, defined as:
\begin{IEEEeqnarray}{rCl}
J_{\mathrm{boundary}} &=& \sum_{i} \sum_{t=1}^{T} \Big[ \max(0, y_i(t) - y_{\mathrm{upper}}^{(i)} - \delta_{\mathrm{boundary}}) \nonumber\\
&& {} + \max(0, y_{\mathrm{lower}}^{(i)} + \delta_{\mathrm{boundary}} - y_i(t)) \Big]
\end{IEEEeqnarray}
This term is zero when vehicles are within the safe zone $[y_{\mathrm{lower}}^{(i)} + \delta_{\mathrm{boundary}}, y_{\mathrm{upper}}^{(i)} - \delta_{\mathrm{boundary}}]$; once vehicles exceed this range, penalty grows linearly with deviation, encouraging vehicles to remain within their drivable regions.

Based on the above anchor and rule consistency joint guidance, we summarize the Stage-2 denoising regeneration process in Algorithm~\ref{alg:stage2}:

\begin{algorithm}[!t]
\caption{Stage-2 Anchor-conditioned Multi-objective Guided Denoising}
\label{alg:stage2}
\begin{algorithmic}[1]
\REQUIRE $K$, $B$, $\alpha$, $R$, $M$, $\epsilon_\theta$, $\operatorname{E}_{\mathrm{LLM}}$, $\{\beta_k, \Sigma_k\}$, $(\lambda_{\mathrm{anchor}}, \lambda_{\mathrm{avoid}}, \lambda_{\mathrm{boundary}})$
\ENSURE $\tau_0$
\STATE $A \gets \operatorname{E}_{\mathrm{LLM}}(R)$
\STATE $\tau_K \sim \mathcal{N}(0, I)$
\FOR{$k = K$ \textbf{down to} $1$}
    \STATE $\hat{\epsilon}_\theta \gets \epsilon_\theta(\tau_k, k, c)$
    \STATE $\hat{\tau}_0 \gets \frac{1}{\sqrt{\bar{\alpha}_k}}(\tau_k - \sqrt{1-\bar{\alpha}_k} \hat{\epsilon}_\theta)$
    \STATE $J_{\mathrm{anchor}} \gets J_{\mathrm{anchor}}(\hat{\tau}_0; A)$
    \STATE $J_{\mathrm{avoid}} \gets J_{\mathrm{avoid}}(\hat{\tau}_0; B)$
    \STATE $J_{\mathrm{boundary}} \gets J_{\mathrm{boundary}}(\hat{\tau}_0; M)$
    \STATE $J \gets \lambda_{\mathrm{anchor}} J_{\mathrm{anchor}} + \lambda_{\mathrm{avoid}} J_{\mathrm{avoid}} + \lambda_{\mathrm{boundary}} J_{\mathrm{boundary}}$
    \STATE $\tau_{k-1} \gets \hat{\tau}_0 - \alpha \Sigma_k \nabla_{\tau_k} J(\hat{\tau}_0)$
\ENDFOR
\RETURN $\tau_0$
\end{algorithmic}
\end{algorithm}

Through the above multi-objective guidance, second-stage generation results improve upon the first stage in three aspects: First, scenario semantics are preserved---collision types and interaction logic described in user instructions are retained; Second, background vehicles gain interactive response capability---the diffusion model learns multi-vehicle interaction patterns from real driving data, enabling background vehicles to adaptively respond with reasonable reactions such as deceleration and yielding based on ego and adversarial vehicle behaviors, rather than simply following original trajectories; Third, trajectory smoothness is improved---velocity and acceleration changes become more continuous, closer to real driving behaviors, with kinematic characteristics better matching original dataset distributions. Thus, the two subtasks are ultimately completed: first solving the controllable scenario generation problem, then solving the realism problem.

% ============================================================
% 4. Experiment
% ============================================================
\section{Experiment}

% Place figure* early to ensure it appears on page 8 top
\begin{figure*}[!t]
\centering
\includegraphics[width=\textwidth]{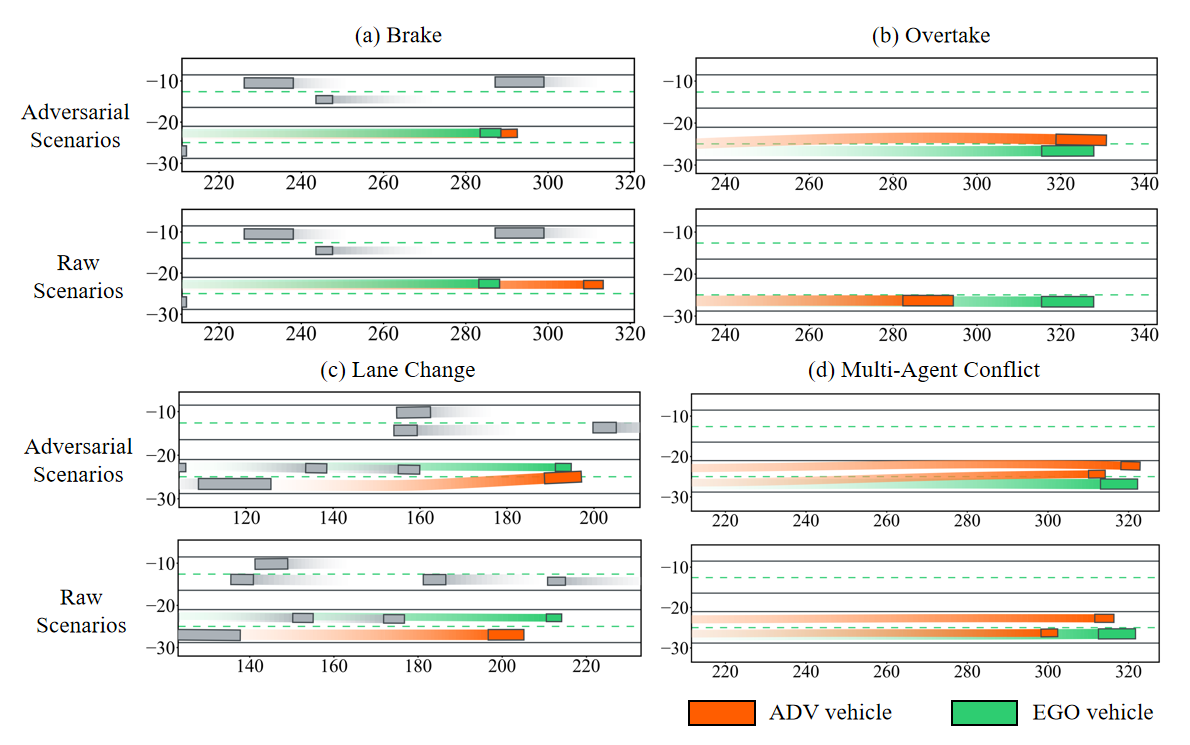}
\caption{Adversarial safety-critical scenarios generated by AnchorDrive. Examples include brake-induced rear-end collision, overtake-and-cut-back collision, dangerous cut-in side collision, and multi-vehicle coordinated conflict, demonstrating the model's capability to generate diverse safety-critical scenarios.}
\label{fig:generated_scenarios}
\end{figure*}

This section presents the experimental evaluation of the proposed AnchorDrive model. We first describe the dataset, evaluation metrics, and experimental setup. Then, we compare AnchorDrive's performance against baseline methods and provide quantitative analysis of the results. Additionally, we conduct ablation studies to evaluate the contribution of key components in our approach.

\subsection{Dataset}

We conduct experiments on the highD dataset. highD is a highway naturalistic driving dataset collected via drone aerial photography, covering 6 recording locations with each road segment approximately 410~m in length; total data duration is approximately 16.5 hours, containing approximately 110,000 vehicle trajectories at 25~Hz sampling frequency. We extract trajectory segments using fixed time windows for model training and evaluation: using 1.28 seconds (32 frames) of past motion data to predict 5.12 seconds (128 frames) of future trajectories.

\subsection{Evaluation Metrics}

To comprehensively evaluate generated scenario quality, we design metrics from three aspects: criticality, realism, and semantic controllability.

\textbf{Criticality}: This aspect measures the effectiveness of generated scenarios. The ego-adversarial collision rate quantifies the percentage of scenarios where adversarial vehicles collide with ego vehicles; higher values indicate stronger adversarial effects.

\textbf{Realism}: To ensure generated scenarios conform to real-world traffic behaviors, we evaluate off-road rates, non-target collision rates, and trajectory distribution consistency. Ego off-road rate, adversarial off-road rate, and background off-road rate measure how frequently ego, adversarial, and background vehicles drive off the road respectively; lower values indicate more reasonable behaviors. Non-target collision rates include ego-background collision rate, adversarial-background collision rate, and background-background collision rate, quantifying collision frequencies between different vehicle groups; lower values indicate interactions more consistent with reality.

Additionally, we compute the first-order Wasserstein distance~\cite{kantorovitch1958translocation} (WD) between generated and real trajectories on velocity and acceleration statistical distributions to measure consistency between generation results and real driving data. Specifically, for any one-dimensional statistic (velocity or acceleration), let the real sample set be $\{x_i\}_{i=1}^n$ and generated sample set be $\{y_i\}_{i=1}^n$, then:
\begin{IEEEeqnarray}{c}
\hat{W}_1(P,Q) = \frac{1}{n}\sum_{i=1}^{n}|x_{(i)} - y_{(i)}|
\end{IEEEeqnarray}
where $x_{(i)}$ and $y_{(i)}$ are sorted sample values respectively. We merge velocity samples from ego and adversarial vehicles to compute velocity WD, merge acceleration samples from both vehicles to compute acceleration WD, and take the average of velocity WD and acceleration WD as the final metric. Smaller values indicate generated trajectories are kinematically closer to real traffic behaviors in statistical distribution.

\textbf{Semantic Controllability}: We use task success rate to measure whether scenarios satisfy description objectives in natural language instructions; higher values indicate greater semantic controllability.

% Place fig4 early to ensure proper page placement
\begin{figure*}[!t]
\centering
\includegraphics[width=\textwidth]{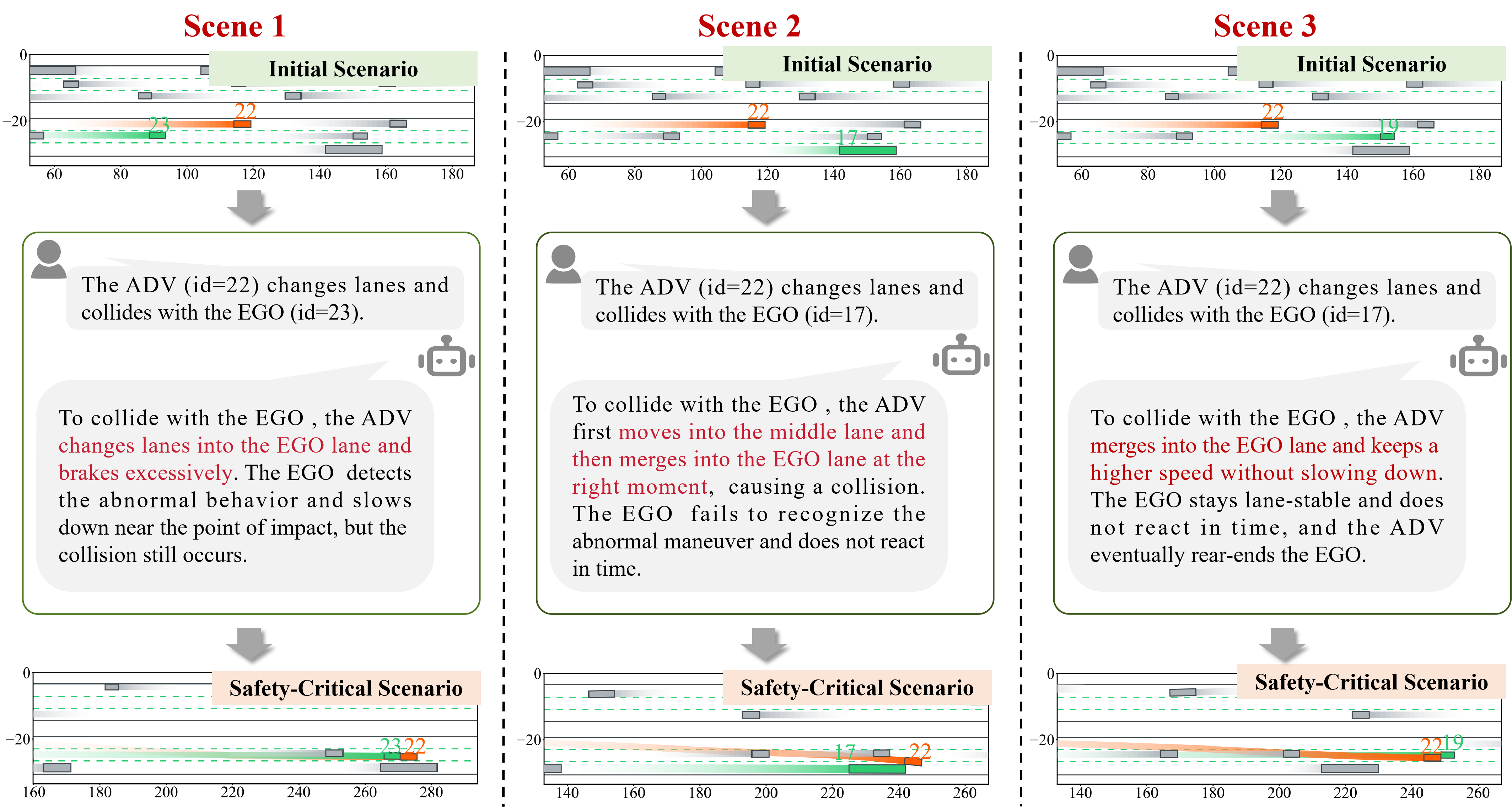}
\caption{Controllability study. Under the same initial scenario conditions, the ADV vehicle is commanded to attack different vehicles, and ADV can make different decisions based on different targets.}
\label{fig:controllability}
\end{figure*}

\subsection{Experimental Setup}

\textbf{Baselines}. We compare our proposed method against existing LLM-based safety-critical scenario generation baseline methods. Specifically, we re-implement and compare with the following two methods: LLMscenario~\cite{chang2024llmscenario}, which sends original complete scenarios with prompts to the LLM, commanding the LLM to modify trajectories in original scenarios to obtain target dangerous scenarios; and LD-scene~\cite{peng2025ld}, which has the LLM generate loss functions based on its understanding of scenarios and user instructions, using these loss functions to guide the denoising process of trained diffusion models to obtain user-specified dangerous scenarios. Both methods are re-implemented by us.

\textbf{Implementation Details}. Our method is implemented in PyTorch and trained for approximately 8 hours on an NVIDIA GeForce RTX 4090 GPU. The trajectory diffusion model is trained for 20 epochs using the Adam optimizer~\cite{kingma2014adam} with learning rate $\mathrm{lr}=0.003$ and 50 denoising steps. The first-stage driver agent uses Gemini-2.5-Pro~\cite{comanici2025gemini} as the decision model, with each planning step length set to 10 frames; second-stage anchor extraction is also performed by Gemini-2.5-Pro.

\subsection{Overall Performance}

As shown in Fig.~\ref{fig:generated_scenarios}, our method can generate diverse traffic scenarios. The figure displays four types of safety-critical scenarios, encompassing both single-vehicle conflict and multi-vehicle coordinated conflict patterns. The corresponding prompts for the four scenarios are: a) ADV suddenly brakes causing EGO to rear-end; b) ADV attempts to overtake EGO but returns to original lane too early, causing collision; c) ADV changes lanes into EGO's lane, colliding with EGO; d) ADV1 (lower) attempts to cut in front of ADV2 (upper), ADV2 fails to decelerate in time causing rapid gap reduction, forcing ADV1 to urgently cut back to original lane to avoid rear-end from ADV2, ultimately colliding with EGO.

% Place table1 early to ensure proper page placement
\begin{table*}[!t]
\centering
\caption{Overall performance comparison with baseline methods on the highD dataset. We compare our proposed method with LLMscenario and LD-scene to evaluate overall performance on criticality, realism, and semantic controllability in safety-critical scenario generation.}
\label{tab:main_results}
\renewcommand{\arraystretch}{1.3}
\resizebox{\textwidth}{!}{%
\begin{tabular}{l c ccccc c}
\toprule
\multirow{2}{*}{\textbf{Method}} & \textbf{Criticality} & \multicolumn{5}{c}{\textbf{Realism}} & \textbf{Controllability} \\
\cmidrule(lr){2-2} \cmidrule(lr){3-7} \cmidrule(lr){8-8}
& \makecell{EGO-ADV\\Coll $\uparrow$} & \makecell{EGO/ADV-BG\\Coll $\downarrow$} & \makecell{BG-BG\\Coll $\downarrow$} & \makecell{EGO/ADV\\Offroad $\downarrow$} & \makecell{BG\\Offroad $\downarrow$} & WD $\downarrow$ & \makecell{Task\\Success $\uparrow$} \\
\midrule
LLMscenario & 0.83 & 0.23 & / & 0.14 & / & 7.64 & 0.78 \\
LD-scene & 0.69 & 0.02 & \textbf{0} & 0.06 & \textbf{0.04} & \textbf{0.72} & 0.55 \\
\textbf{AnchorDrive (Ours)} & \textbf{0.86} & \textbf{0} & \textbf{0} & \textbf{0.02} & \textbf{0.04} & 1.15 & \textbf{0.81} \\
\bottomrule
\end{tabular}%
}
\end{table*}

Table~\ref{tab:main_results} presents quantitative comparison results between our method and baselines. Overall, the proposed AnchorDrive demonstrates stronger comprehensive advantages across criticality, realism, and semantic controllability metrics. Specifically, AnchorDrive's EGO--ADV Coll is 0.86, higher than LLMscenario (0.83) and LD-scene (0.69), indicating our method more easily triggers target adversarial collisions; meanwhile, AnchorDrive's Task Success reaches 0.81, superior to LLMscenario (0.78) and LD-scene (0.55), demonstrating more stable instruction consistency.

Regarding realism-related metrics, AnchorDrive's EGO/ADV Offroad is 0.02, lower than LLMscenario (0.14) and LD-scene (0.06); additionally, AnchorDrive's EGO/ADV--BG Coll and BG--BG Coll are both 0, showing more robust overall performance. For distribution consistency, LD-scene has the lowest Wasserstein Distance (0.72), while AnchorDrive is 1.15, better than LLMscenario (7.64), indicating our method maintains good trajectory distribution consistency while achieving high criticality and task success rates. In summary, AnchorDrive achieves a better balance among criticality, realism, and semantic controllability.

\subsection{Controllability Study}

This section further evaluates AnchorDrive's responsiveness to natural language instructions. We validate its semantic controllability through the following approach: changing prompts under the same initial scenario and observing whether generated scenario interaction patterns change with semantic variations. Fig.~\ref{fig:controllability} presents corresponding visualization results.

As shown in Fig.~\ref{fig:controllability}, we command the ADV vehicle to attack different vehicles by simply modifying prompts to specify different target vehicle IDs. Results show that the model can stably map the ``target vehicle ID'' semantic constraint to generated behaviors: when targets change, ADV's key decisions adjust accordingly.

% Place fig5 and fig6 early to ensure proper page placement
\begin{figure}[!t]
\centering
\includegraphics[width=\linewidth]{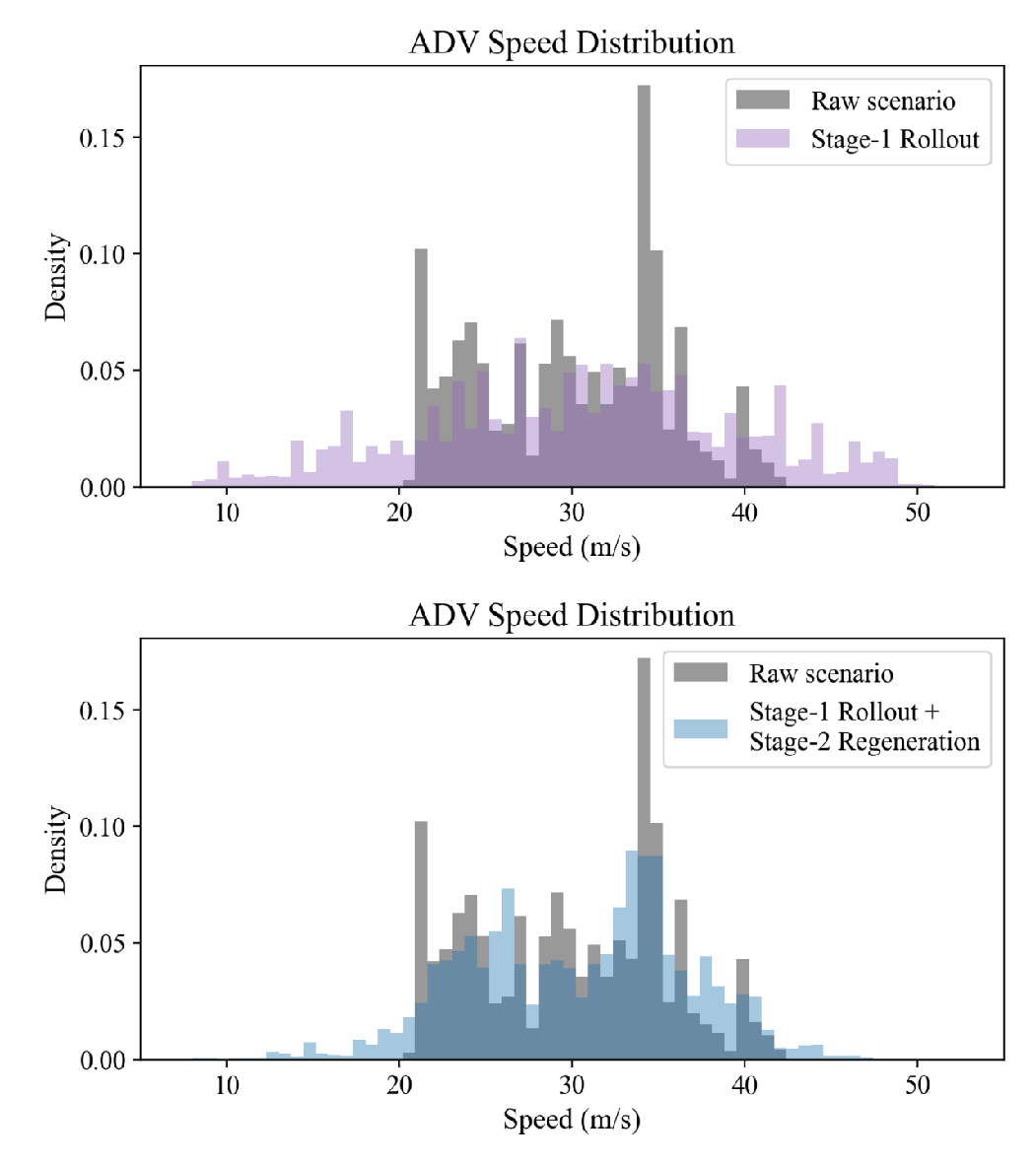}
\caption{Comparison of ADV velocity distributions before and after trajectory regeneration. Comparing adversarial vehicle (ADV) velocity histograms between original scenarios and different generation configurations, velocity distributions become closer to original scenarios after adding Stage-2 trajectory regeneration.}
\label{fig:velocity_distribution}
\end{figure}

\begin{figure*}[!t]
\centering
\includegraphics[width=\textwidth]{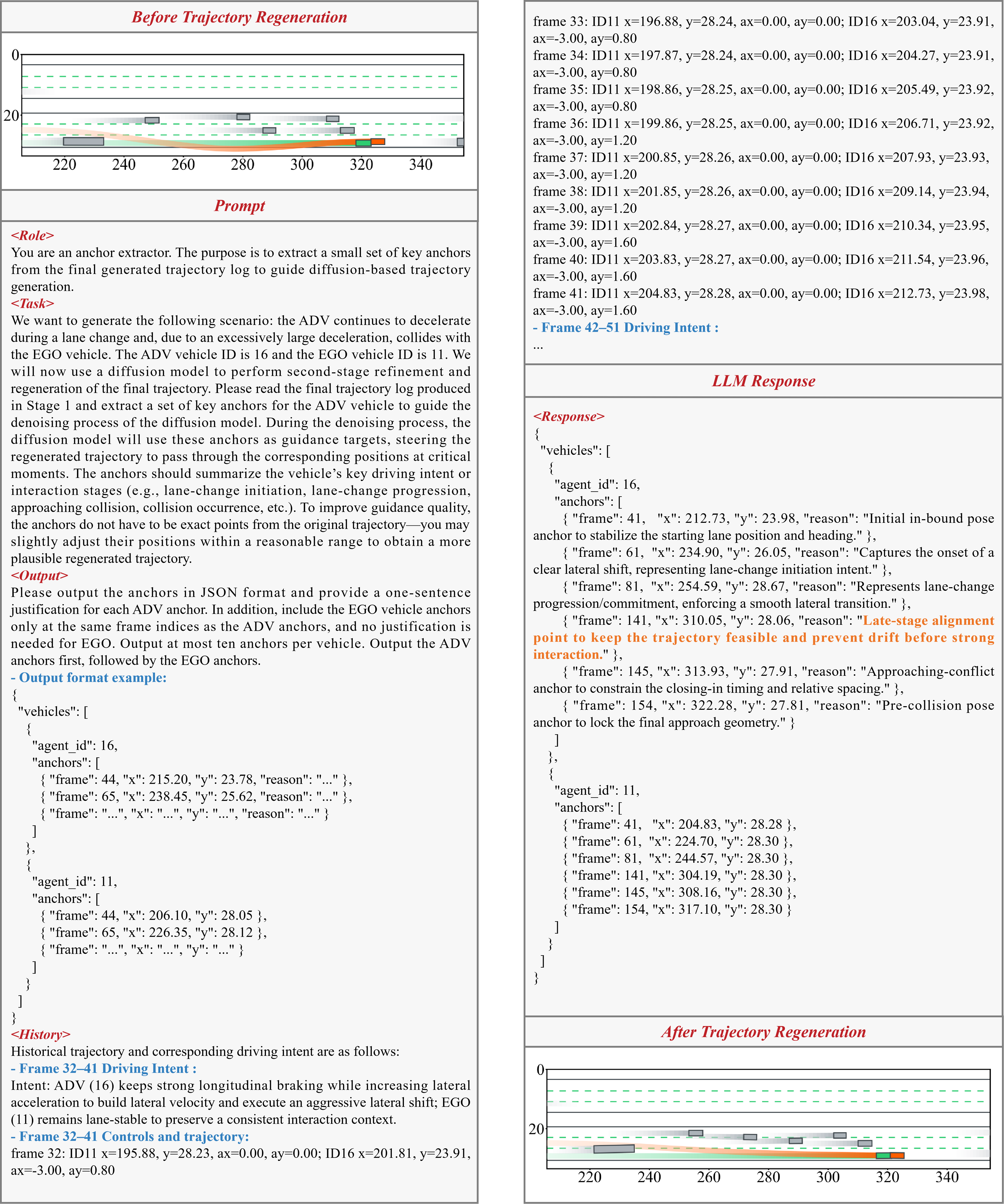}
\caption{Anchor guidance study. This figure demonstrates the role of the LLM-based anchor extractor in second-stage trajectory regeneration. By extracting anchors through the LLM-based anchor extractor, the diffusion model can suppress boundary violations and other unreasonable behaviors while preserving scenario semantics.}
\label{fig:anchor_guidance}
\end{figure*}

\subsection{Ablation Study}

This section evaluates the contribution of key components in AnchorDrive to generation quality. We focus on the first-stage plan assessor decision review mechanism and the second-stage trajectory regeneration module, constructing three configurations for comparison. Table~\ref{tab:ablation} summarizes quantitative results for the three configurations.

\begin{table*}[!t]
\centering
\caption{Ablation study of key components. Comparison of different module configurations (plan assessor decision review and Stage-2 trajectory regeneration) on safety-critical scenario generation performance on the highD dataset.}
\label{tab:ablation}
\renewcommand{\arraystretch}{1.3}
\resizebox{\textwidth}{!}{%
\begin{tabular}{cc c ccccc c}
\toprule
\multicolumn{2}{c}{\textbf{Configuration}} & \textbf{Criticality} & \multicolumn{5}{c}{\textbf{Realism}} & \textbf{Controllability} \\
\cmidrule(lr){1-2} \cmidrule(lr){3-3} \cmidrule(lr){4-8} \cmidrule(lr){9-9}
\makecell{Plan\\Assessor} & \makecell{Traj.\\Regen.} & \makecell{EGO-ADV\\Coll $\uparrow$} & \makecell{EGO/ADV-BG\\Coll $\downarrow$} & \makecell{BG-BG\\Coll $\downarrow$} & \makecell{EGO/ADV\\Offroad $\downarrow$} & \makecell{BG\\Offroad $\downarrow$} & WD $\downarrow$ & \makecell{Task\\Success $\uparrow$} \\
\midrule
$\times$ & $\times$ & 0.85 & 0.15 & / & 0.07 & / & 2.18 & 0.72 \\
$\checkmark$ & $\times$ & \textbf{0.91} & 0.06 & / & 0.05 & / & 2.43 & \textbf{0.85} \\
$\checkmark$ & $\checkmark$ & 0.86 & \textbf{0} & \textbf{0} & \textbf{0.02} & \textbf{0.04} & \textbf{1.15} & 0.81 \\
\bottomrule
\end{tabular}%
}
\end{table*}

First, the plan assessor decision review mechanism significantly improves stability and instruction achievement rate of first-stage closed-loop generation. Compared to configurations without plan assessor, introducing plan assessor raises EGO--ADV coll from 0.85 to 0.91 and Task Success from 0.72 to 0.85, while reducing EGO/ADV--BG coll from 0.15 to 0.06, demonstrating that review and failure feedback can effectively suppress non-target conflicts and boundary violations.

On the other hand, second-stage trajectory regeneration primarily improves trajectory realism and suppresses unreasonable actions. When removing this module under plan assessor-enabled settings, EGO--ADV coll rises from 0.86 to 0.91 and Task Success rises from 0.81 to 0.85, with generation strategy tending toward more aggressive approaches to complete instructions; however, Wasserstein Distance significantly increases from 1.15 to 2.43, EGO/ADV Offroad rises from 0.02 to 0.05, and non-target collision EGO/ADV--BG coll rises from 0.00 to 0.06, indicating less realistic trajectories. After introducing second-stage trajectory regeneration, trajectory realism improves significantly; although EGO--ADV coll and Task Success decrease somewhat, this is a necessary trade-off to ensure realism. Fig.~\ref{fig:velocity_distribution} compares ADV vehicle velocity distributions before and after diffusion optimization; post-optimization distributions are more consistent with original scenarios, with correspondingly smaller Wasserstein Distances.

The above experiments validate two key designs of AnchorDrive: the first-stage plan assessor improves instruction execution stability and interaction reasonableness, while second-stage diffusion optimization significantly enhances trajectory realism.

\subsection{Anchor Guidance Study}

Fig.~\ref{fig:anchor_guidance} demonstrates the necessity and effectiveness of introducing the LLM-based anchor extractor in second-stage refinement. The upper portion of the figure shows safety-critical scenarios directly generated by the first-stage LLM-based driver agent. It can be seen that while this method can achieve instruction-required target interactions, errors may still occur during closed-loop generation; for example, in the figure, ADV's longitudinal velocity is too high during lane change and fails to apply reverse acceleration in time, causing slight boundary violation.

It can be observed that LLM-extracted anchors can summarize ADV's key intentions at each phase, and after detecting lane-change boundary violation segments, proactively skip abnormal segments, setting final alignment anchors before strong interaction occurs to pull the vehicle back to reasonable positions within the lane. Ultimately, under anchor guidance, trajectories generated by the diffusion model both maintain original driving intentions and interaction structures while effectively suppressing boundary violations and other unreasonable behaviors, improving trajectory realism.

% ============================================================
% 5. Conclusion
% ============================================================
\section{Conclusion}

We propose AnchorDrive, a two-stage safety-critical scenario generation framework that simultaneously achieves semantic controllability and trajectory realism. In the first stage, we deploy an LLM as a driver agent within a simulation closed-loop, reasoning over scenario states under user natural language instruction constraints and progressively outputting control commands to generate high-risk interaction processes conforming to instructions. In the second stage, we utilize the LLM to extract key anchor points from first-stage trajectories and use anchors as primary guidance objectives, jointly with other guidance terms to guide the diffusion model's denoising process for trajectory regeneration, improving generated trajectory realism while preserving instruction semantic objectives. Experimental results on the highD dataset demonstrate that the AnchorDrive framework can stably generate safety-critical scenarios consistent with user instruction semantics on one hand, while making generated trajectories statistically closer to real data distributions on the other, achieving a balance between controllability and realism.

Our method still has room for further improvement. In the future, we plan to conduct the following work to further enhance the framework's capabilities: 1) Current experiments are primarily validated on highD; subsequent work will extend to more complex urban scenario datasets (such as nuScenes) to cover richer road topologies, traffic participant types, and interaction patterns, further examining the method's generalization capability; 2) The current framework's scenario generation speed is relatively slow, primarily because the first stage requires multiple rounds of closed-loop reasoning and plan assessor review iterations, and the second stage additionally requires diffusion model trajectory regeneration, resulting in high overall computational overhead; subsequent research can explore methods to improve generation efficiency.

% References
\bibliographystyle{IEEEtran}
\bibliography{assets/references}

\end{document}